\documentclass[journal=jacsat,manuscript=article]{achemso}
\usepackage{graphicx}
\usepackage{hyperref}
\usepackage{xcolor}
\usepackage[subrefformat=parens]{subcaption}
\usepackage[frozencache,cachedir=.]{minted}

\usepackage[version=3]{mhchem} 
\author{Jocelyn Sunseri}
\author{David R. Koes}
\affiliation{Department of Computational and Sytems Biology, University of Pittsburgh}
\email{dkoes@pitt.edu}

\title{libmolgrid: GPU Accelerated Molecular Gridding for Deep Learning Applications}

\keywords{Machine Learning, Deep Learning, Molecular Modeling, Python}
\begin{document}
\begin{abstract}
There are many ways to represent a molecule as input to a machine learning model and each is associated with loss and retention of certain kinds of information. In the interest of preserving three-dimensional spatial information, including bond angles and torsions, we have developed \texttt{libmolgrid}, a general-purpose library for representing three-dimensional molecules using multidimensional arrays. This library also provides functionality for composing batches of data suited to machine learning workflows, including data augmentation, class balancing, and example stratification according to a regression variable or data subgroup, and it further supports temporal and spatial recurrences over that data to facilitate work with recurrent neural networks, dynamical data, and size extensive modeling. It was designed for seamless integration with popular deep learning frameworks, including Caffe, PyTorch, and Keras, providing good performance by leveraging graphical processing units (GPUs) for computationally-intensive tasks and efficient memory usage through the use of memory views over preallocated buffers. \texttt{libmolgrid} is a free and open source project that is actively supported, serving the growing need in the molecular modeling community for tools that streamline the process of data ingestion, representation construction, and principled machine learning model development. 
\end{abstract}

\section{Introduction}
The field of computational chemistry has grown in tandem with computing resources\cite{chemistry1995mathematical, mattson1995parallel, leach2001molecular, council2003beyond, cramer2004essentials} and quantitative data about molecular structure and thermodynamics\cite{berman2002protein, gaulton2011chembl, yeguas2014big, kim2015pubchem, goh2017deep, hu2017entering}. In particular, machine learning has emerged as a novel area of study that holds great promise for unprecedented improvements in predictive capabilities for such problems as virtual screening\cite{jorissen2005}, binding affinity prediction\cite{Ballester2010, zilian2013sfcscore}, pose prediction\cite{Ashtawy2015, chupakhin2013predicting}, and lead optimization\cite{ekins2014combining, yasuo2018predicting, zhou2019optimization, jimenez2019deltadelta}. The representation of input data can fundamentally limit or enhance the performance and applicability of machine learning algorithms\cite{lusci2013deep, duvenaud2015convolutional, gomez2018automatic}. Standard approaches to data representation include performing initial feature selection based on various types of molecular descriptors/fingerprints\cite{xue2004prediction, todeschini2008handbook, unterthiner2015toxicity}, including simple molecular properties\cite{ballester2014does, kundu2018machine}, molecular connectivity and shape\cite{cruciani2000volsurf, skalic2019shape}, electro-topological state\cite{katritzky1993traditional, durrant2011binana, kier1999molecular, cang2017topologynet}, quantum chemical properties\cite{karelson1996quantum}, and geometrical properties\cite{katritzky1993traditional} (or a combination of multiple of these descriptor categories\cite{durrant2011nnscore, wu2018moleculenet}); summarizing inputs using representations that are amenable to direct algorithmic analysis while preserving as much relevant information as possible, such as pairwise distances between all or selected atom groups\cite{deng2004predicting, Ballester2010, gomes2017atomic,  schutt2017schnet}, using Coulomb matrices or representations derived from them\cite{hansen2013assessment, hansen2015machine}, or encoding information about local atomic environments that comprise a molecule\cite{behler2011atom, bartok2013representing, smith2017ani}; or using some representation of molecular structure directly as input to a machine learning algorithm such as a neural network\cite{ragoza2017protein,  wallach2015atomnet, duvenaud2015convolutional, schutt2017moleculenet, gomes2017atomic, kdeep}, which extracts features and creates an internal representation of molecules itself as part of training. Commonly used input representations for the latter method include SMILES and/or InChi strings\cite{gomez2018automatic, winter2019learning}, molecular graphs\cite{lusci2013deep, urban2018inner, duvenaud2015convolutional, kearnes2016molecular, pham2018graph, feinberg2018potentialnet}, and voxelized spatial grids\cite{wallach2015atomnet, ragoza2017protein} representing the locations of atoms. 

Among this latter form of molecular representation, spatial grids possess certain virtues including minimal overt featurization by the user (theoretically permitting greater model expressiveness) and full representation of three-dimensional spatial interactions in the input. For regular cubic grids, this comes at the cost of coordinate frame dependence, which can be ameliorated by data augmentation and can also be theoretically addressed with various types of inherently equivariant network architectures or by using other types of multidimensional grids. Spatial grids have been applied successfully to tasks relevant to computational chemistry like virtual screening\cite{wallach2015atomnet, ragoza2017protein, skalic2018playmolecule}, pharmacophore generation\cite{skalic2018ligvoxel}, molecular property prediction\cite{kajita2017universal, kdeep}, molecular classification\cite{amidi2018enzynet, kajita2017universal}, protein binding site prediction\cite{hendlich1997ligsite, jimenez2017deepsite, jiang2019novel}, molecular autoencoding\cite{kuzminykh20183d}, and generative modeling\cite{brock2016generative, biggan, thomas2018tensor} by both academic and industrial groups, demonstrating their general utility. 

Chemical datasets have many physical and statistical properties that prove problematic for machine learning approaches and special care must be taken to manage them. Classes are typically highly imbalanced in available datasets, with many more known inactive than active compounds for a given protein target; regression tasks may span many orders of magnitude, with nonuniform representation of the underlying chemical space at particular ranges of the regressor; and examples with matching class labels or regression target values may also be unequally sampled from other underlying classes (e.g. there may be significantly more binding affinity data available for specific proteins that have been the subject of greater medical attention, such as the estrogen receptors, or for protein classes like kinases). Chemical space is characterized by inherent symmetries that may not be reflected in a given molecular representation format. The pathologies unique to cubic grids were already mentioned, but in general all available representation methods require tradeoffs among the desired goals of maintaining symmetry to translation, rotation, and permutation while also preserving all the relevant information about chemical properties and interactions. Computational efficiency must also be prioritized if the final method is to have practical use. End users should consider the required tradeoffs and make a choice about input representation informed by their application, but once the choice is made, provided they have chosen a common representation format, the speed, accuracy, and reproducibility of their work will be enhanced if they can use a validated, open source library for input generation. By offloading data processing tasks commonly required for machine learning workflows to an open source library specialized for chemical data, computational chemists can systematically obtain better results in a transparent manner. 

Using multidimensional grids (i.e. voxels) to represent atomic locations (and potentially distributions) is computationally efficient - their generation is embarrassingly parallel and therefore readily amenable to modern GPU architectures - and preserves three dimensional spatial relationships present in the original input. Their coordinate frame dependence can be removed or circumvented. But commonly available molecular parsing and conversion libraries do not yet provide gridding functionality; nor do they implement the other tasks a data scientist would require to obtain good performance on typical chemical datasets, such as the strategic resampling and data augmentation routines detailed above. Thus we abstracted the gridding and batch preparation functionality from our past work, \texttt{gnina}\cite{ragoza2017protein}, into a library that can be used for general molecular modeling tasks but also interfaces naturally with popular Python deep learning libraries. Implemented in C++ with Python bindings, \texttt{libmolgrid} is a free and open source project intended to democratize access to molecular modeling via multidimensional arrays and to provide the additional functionality necessary to get good results from training machine learning models with typical chemical datasets. 

\section{Implementation}
Key \texttt{libmolgrid} functionality is implemented in a modular fashion to ensure maximum versatility. Essential library features are abstracted into separate classes to facilitate use independently or in concert as required by a particular application. 

\subsection{Grids}
\begin{figure}[tbp]
\centering
\begin{subfigure}[t]{.5\linewidth}
\includegraphics[width=\linewidth]{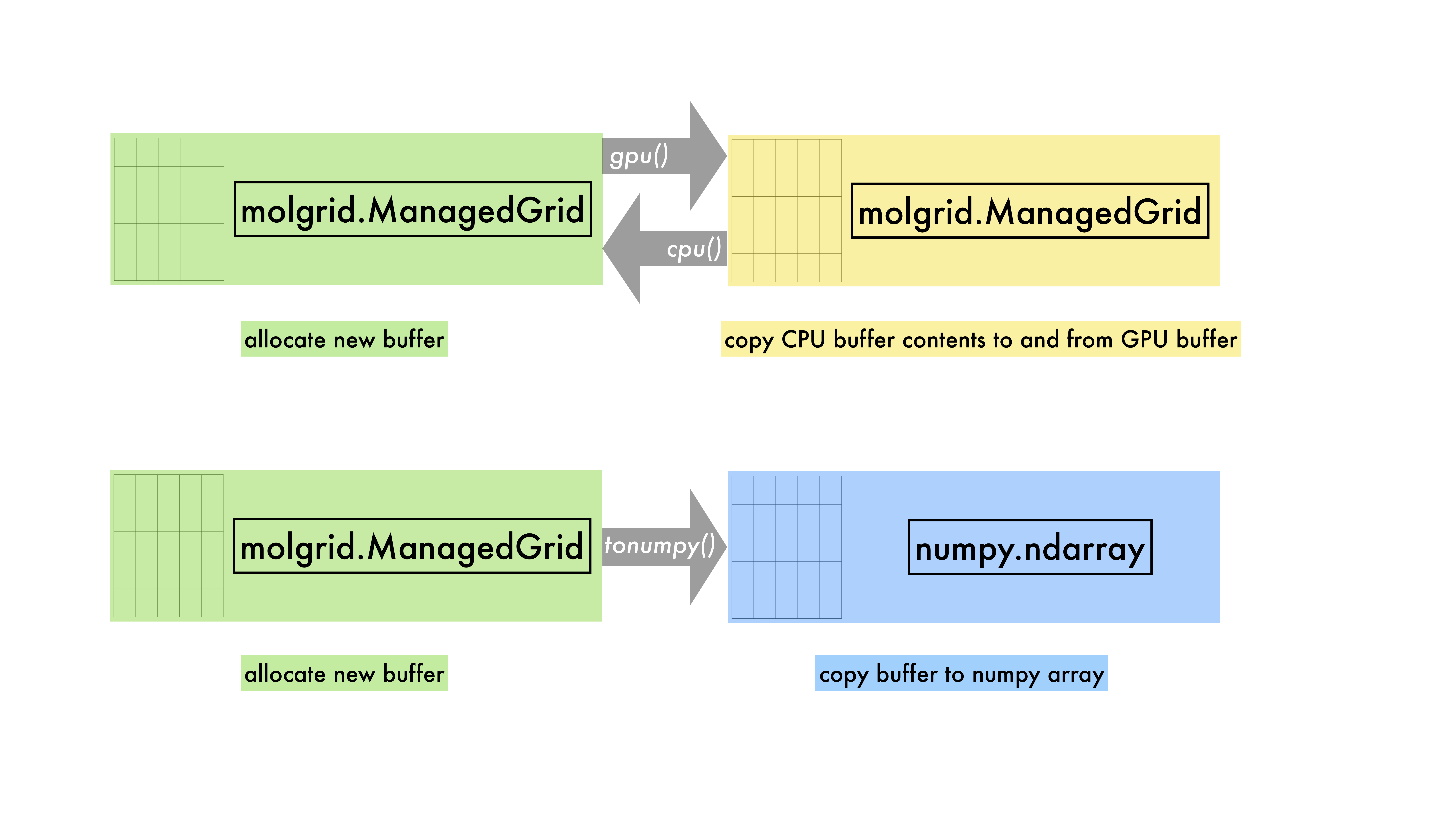}
\caption[]{\label{mgrid}}
\end{subfigure}%
\hfill
\begin{subfigure}[t]{.5\linewidth}
\includegraphics[width=\linewidth]{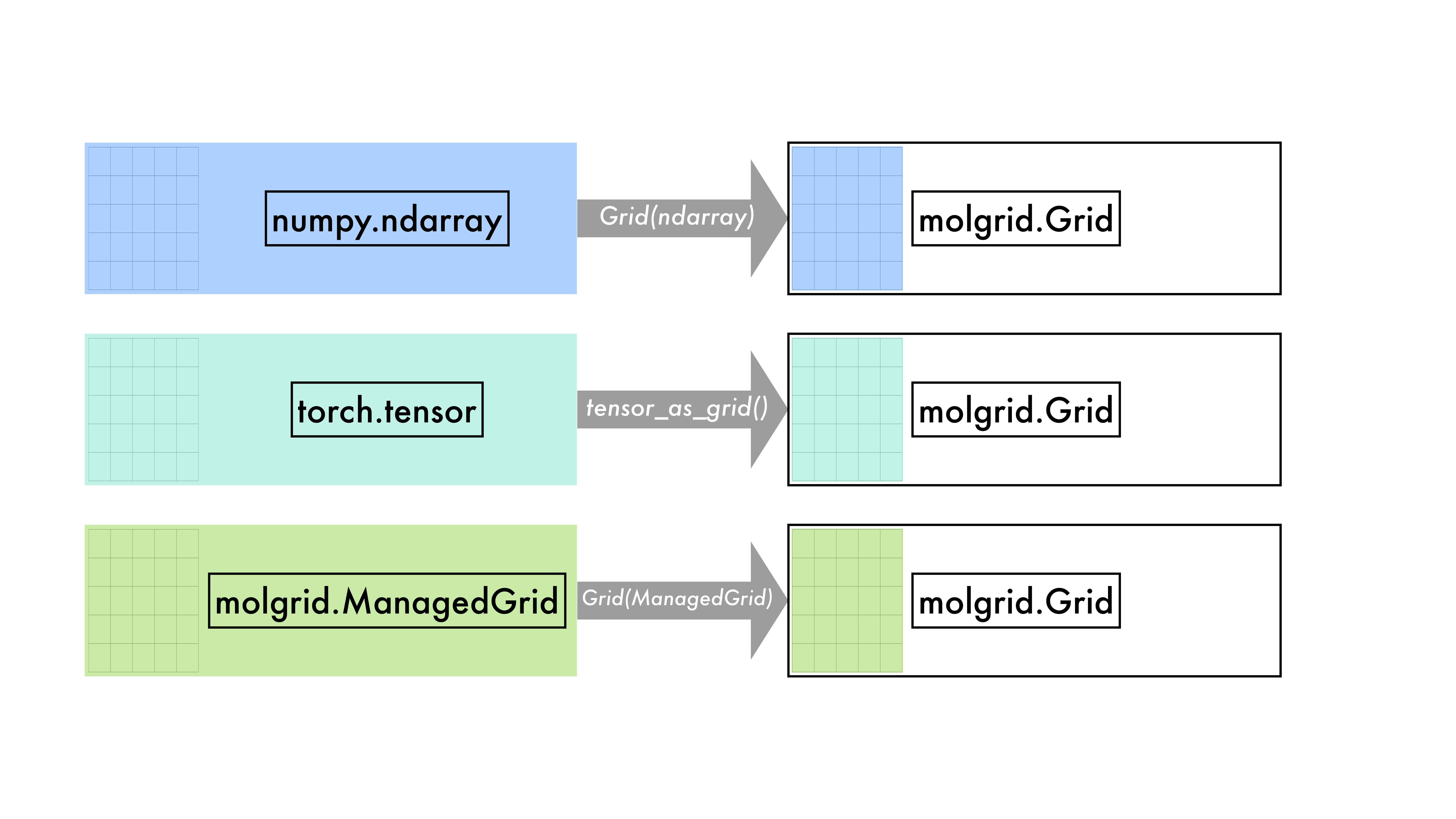}
\caption[]{\label{gridview}}
\end{subfigure}
\caption{\label{fig:grids} \texttt{ManagedGrid}s manage their own memory buffer, which can migrate data between the CPU and GPU and copy data to a NumPy array as shown in \subref{mgrid}. \texttt{Grid}s are a view over a memory buffer owned by another object; they may be constructed from a Torch tensor, a \texttt{ManagedGrid}, or an arbitrary data buffer with a Python-exposed pointer, including a NumPy array as shown in \subref{gridview}.
}
\end{figure}
The fundamental object used to represent data in \texttt{libmolgrid} is a multidimensional array which the API generically refers to as a grid. Grids are typically used during training to represent voxelized input molecules or matrices of atom coordinates and types. They can be constructed in two flavors, \texttt{Grid}s and \texttt{ManagedGrid}s; \texttt{ManagedGrid}s manage their own underlying memory, while \texttt{Grid}s function as views over a preexisting memory buffer. Figure~\ref{fig:grids} illustrates the behavior of \texttt{ManagedGrid}s (\ref{mgrid}) and \texttt{Grid}s (\ref{gridview}). \texttt{ManagedGrid}s can migrate data between devices, and they create a copy when converting to or from other objects that have their own memory. \texttt{Grid}s do not own memory, instead serving as a view over the memory associated with another object that does; they do not create a copy of the buffer, rather they interact with the original buffer directly, and they cannot migrate it between devices. \texttt{Grid}s and \texttt{ManagedGrid}s are convertible to NumPy arrays as well as Torch tensors. 

Because of automatic conversions designed for PyTorch interoperability, a user intending to leverage basic batch sampling, grid generating, and transformation capabilities provided by \texttt{libmolgrid} in tandem with PyTorch for neural network training can simply use Torch tensors directly, with little to no need for explicit invocation of or interaction with \texttt{libmolgrid} grids. Memory allocated on a GPU via a Torch tensor will remain there, with grids generated in-place. An example of this type of usage is shown in the first example in Listing~\ref{lst:grid}. 

A \texttt{Grid} may also be constructed explicitly from a Torch tensor, a NumPy array, or if necessary from a pointer to a memory buffer. Examples of constructing a \texttt{Grid} from a Torch tensor are shown in the second usage section in  Listing~\ref{lst:grid}. The third usage section shows provided functionality for copying NumPy array data to \texttt{ManagedGrid}s, while the fourth usage section shows functionality for constructing \texttt{Grid} views over NumPy array data buffers. In the fourth example, note that in recent NumPy versions the default floating-point data type is float64, and therefore the user must specify float32 as the dtype if intending to copy the array data into a float rather than a double \texttt{Grid}. 

\begin{listing}[]
\begin{minted}[
frame=single,
framesep=2mm,
baselinestretch=1.2,
fontsize=\footnotesize]{python}
# Usage 1: molgrid functions taking Grid objects can be passed Torch tensors directly, 
# with conversions managed internally
tensor = torch.zeros(tensor_shape, dtype=torch.float32, device='cuda')
molgrid.gmaker.forward(batch, input_tensor)

# Usage 2: construct Grid as a view over a Torch tensor with provided helper function
tensor = torch.zeros((2,2), dtype=torch.float32, device='cuda')
grid = molgrid.tensor_as_grid(tensor) # dimensions and data location are inferred
# alternatively, construct Grid view over Torch tensor directly
grid = molgrid.Grid2fCUDA(tensor)

# Usage 3: copy ManagedGrid data to NumPy array
# first, construct a ManagedGrid
mgrid = molgrid.MGrid1f(batch_size)
# copy to GPU and do work on it there
mgrid.gpu()
# (do work)
# copy ManagedGrid data to a NumPy array with helper function;
# this copies data back to the CPU if necessary
array1 = mgrid.tonumpy() 
# alternatively, construct NumPy array with a copy of ManagedGrid CPU data;
# must sync to CPU first
mgrid.cpu()
array2 = np.array(mgrid)

# Usage 4: construct Grid from NumPy array
array3 = np.zeros((2,2), dtype=np.float32) # must match source and destination dtypes
tensor = molgrid.Grid2f(array3)
\end{minted}
\caption{Examples of \texttt{Grid} and \texttt{ManagedGrid} usage.}
\label{lst:grid}
\end{listing}

The explicit specialization of a grid exposed in the Python \texttt{molgrid} API has a naming convention that specifies its dimensionality, underlying data type, and in the case of \texttt{Grid}s, the device where its memory buffer is located. The structure of the naming convention is \texttt{[GridClass][NumDims][DataType]["CUDA" if GridClass=="Grid" and DataLoc == "GPU"]}. Since \texttt{ManagedGrid}s can migrate their data from host to device, their names do not depend on any particular data location. For example, a 1-dimensional  \texttt{ManagedGrid} of type float is an \texttt{MGrid1f}, a 3-dimensional \texttt{Grid} of type float is a \texttt{Grid3f}, and a 5-dimensional \texttt{Grid} of type double that is a view over device data is a \texttt{Grid5dCUDA}.

\subsection{Atom Typing}

Several atom typing schemes are supported, featuring flexibility in the ways types are assigned and represented. Atoms may be typed according to XS atom typing, atomic element, or a user-provided callback function. Types may be represented by a single integer or a vector encoding. For a typical user, typing (with either index or vector types) can be performed automatically via an \texttt{ExampleProvider}.

\subsection{Examples}

\texttt{Example}s consist of typed coordinates that will be analyzed together, along with their labels. An \texttt{Example} may consist of multiple \texttt{CoordinateSet}s (which may each utilize a different scheme for atom typing) and may be one of a sequence of \texttt{Example}s within a group. For example, a single \texttt{Example} may have a \texttt{CoordinateSet} for a receptor and another \texttt{CoordinateSet} for a ligand to be scored with that receptor, or perhaps multiple \texttt{CoordinateSet}s corresponding to multiple poses of a particular ligand. \texttt{Example}s may be part of a group that will be processed in sequence, for example as input to a recurrent network; in that case distinct groups are identified with a shared integer value, and a sequence continuation flag indicates whether a given \texttt{Example} is a continuation of a previously observed sequence or is initiating a new one. 

\subsection{ExampleProvider}

\begin{figure}[tbp]
\centering
\includegraphics[width=\linewidth]{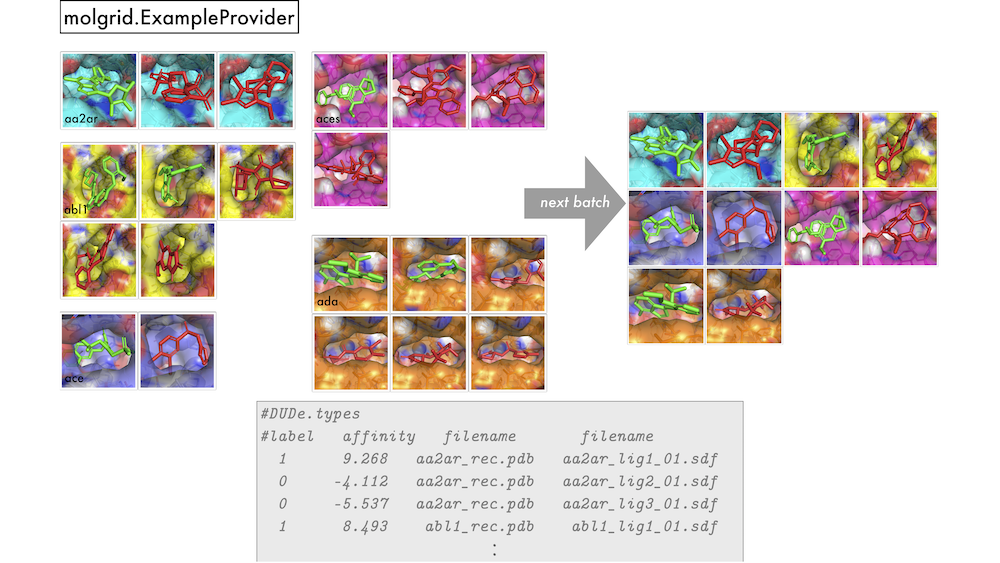}
\begin{minted}[
frame=lines,
framesep=2mm,
baselinestretch=1.2,
fontsize=\footnotesize,
linenos]{python}
exprovider = molgrid.ExampleProvider(shuffle=True, balanced=True, stratify_receptor=True)
exprovider.populate('DUDe.types')
batch = exprovider.next_batch(10)
\end{minted}
\caption{\label{fig:exprovider} An illustration of \texttt{molgrid.ExampleProvider} usage, sampling a batch of 10 randomized, balanced, and receptor-stratified examples from a dataset.} 
\end{figure}

To obtain strategically sampled batches of data for training, a user can employ an \texttt{ExampleProvider} with custom settings. The \texttt{ExampleProvider} constructor is used to create an \texttt{ExampleProvider} with the desired settings, which can then be populated with one or more files specifying examples. The methods \texttt{next} and \texttt{next\_batch} are then used to obtain the next \texttt{Example} or the next batch of \texttt{Example}s, respectively. Figure~\ref{fig:exprovider} shows graphically how an \texttt{ExampleProvider} might obtain a batch of 10 shuffled, class-balanced, receptor-stratified \texttt{Example}s from a larger dataset, with accompanying code. 

Currently, the simplest way to initialize a provider is to populate it with one or more files that specify metadata for \texttt{Example}s, with one \texttt{Example} per line. At a high level, that line will specify class and regression target values for the \texttt{Example}, any group identification associated with the \texttt{Example} (i.e. a shared integer label identifying \texttt{Example}s to be processed sequentially, as with temporal data provided as input to a recurrent network), and then one or more strings identifying filenames of molecules corresponding to that \texttt{Example}. The default line layout is \texttt{[(int)group][(float)label]*[molfile]+}.  An example is shown in Figure~\ref{fig:exprovider}. In the examples we provide with our project, these files have a \texttt{.types} suffix.

Many \texttt{ExampleProvider} options govern \texttt{Example} resampling and the layout of the lines within the metadata file used to populate the \texttt{ExampleProvider} that will be used to determine how that resampling will be performed. Listing~\ref{lst:exp_constructor} shows all the available options at the time of construction. Randomization is enabled with the \texttt{shuffle} option; oversampling of underrepresented classes to provide equal representation from all available classes categorized by the \texttt{Example} label is enabled with \texttt{balanced}; resampling based on a specific molecule associated with an \texttt{Example} (determined by the first filename encountered on a given metadata line) comes from \texttt{stratify\_receptor} (as the name suggests, this is often used to sample equally from \texttt{Example}s associated with different receptors); \texttt{labelpos} specifies the location of the binary classification label on each line of the metadata file, in terms of an index starting from 0 that numbers the entries on a line; \texttt{stratify\_pos} similarly specifies the location of a regression target value that will be used to stratify \texttt{Example}s for resampling (for example a binding affinity); \texttt{stratify\_abs} indicates that stratification of \texttt{Example}s based on a regression value will use the absolute value, which is useful when a negative value has a special meaning such as with a hinge loss; and \texttt{stratify\_min}, \texttt{stratify\_max}, and \texttt{stratify\_step} are used to define the bins for numerical stratification of \texttt{Example}s. 

Additional options provide customization for interpreting examples and optimizations for data I/O. When using a recurrent network for processing a sequence of data, such as the case of training with molecular dynamics frames,  \texttt{group\_batch\_size} specifies the number of frames to propagate gradients through for truncated backpropagation through time and \texttt{max\_group\_size} indicates the total number of \texttt{Example}s associated with the largest \texttt{Example} group (e.g. the maximum number of frames). \texttt{add\_hydrogens} will result in protonation of parsed molecules with OpenBabel. \texttt{duplicate\_first} will clone the first \texttt{CoordinateSet} in an \texttt{Example} to be separately paired with each of the subsequent \texttt{CoordinateSet}s in that \texttt{Example} (e.g., a single receptor structure is replicated to match different ligand poses).  \texttt{num\_copies} emits the same example multiple times (this allows the same structure to be presented to the neural network using multiple transformations in a single batch). \texttt{make\_vector\_types} will represent types as a one-hot vector rather than a single index. \texttt{cache\_structs} will keep coordinates in memory to reduce training time.  \texttt{data\_root} allows the user to specify a shared parent directory for molecular data files, which then allows the metadata file to specify the filenames as relative paths. Finally, \texttt{recmolcache} and \texttt{ligmolcache} are binary files that store an efficient representation of all receptor and ligand files to be used for training, with each structure stored only once.  These are created using the \texttt{create\_caches2.py} script from \url{https://github.com/gnina/scripts}. Caches combine many small files into one memory mapped file resulting in a substantial I/O performance improvement and reduction in memory usage during training.

\begin{listing}[]
\begin{minted}[
frame=single,
framesep=2mm,
baselinestretch=1.2,
fontsize=\footnotesize]{python}
exprovider = molgrid.ExampleProvider(shuffle=False, balanced=False, 
stratify_receptor=False, labelpos=0, stratify_pos=1, stratify_abs=True, stratify_min=0, 
stratify_max=0, stratify_step=0, group_batch_size=1, max_group_size=0, 
cache_structs=True, add_hydrogens=True, duplicate_first=False, num_copies=1, 
make_vector_types=False, data_root="", recmolcache="", ligmolcache="")
\end{minted}
\caption{Available arguments to \texttt{ExampleProvider} constructor, along with their default values.}
\label{lst:exp_constructor}
\end{listing}

\subsection{GridMaker}

\begin{figure}[tbp]
\centering
\includegraphics[width=\linewidth]{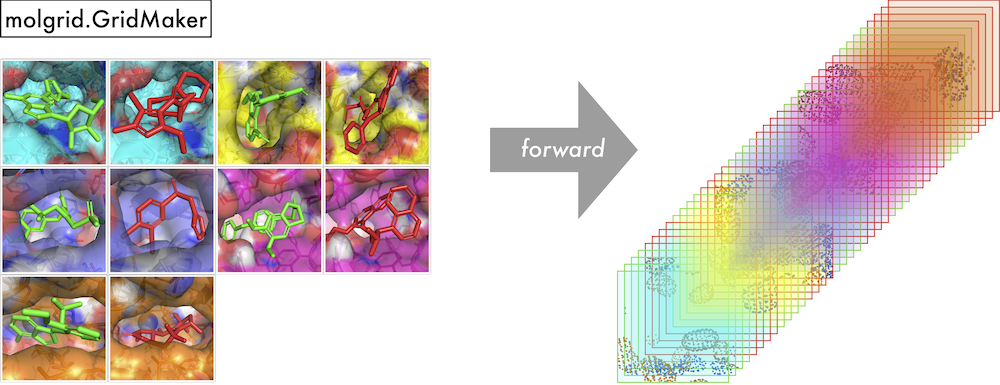}
\begin{minted}[
frame=lines,
framesep=2mm,
baselinestretch=1.2,
fontsize=\footnotesize, linenos]{python}
gmaker = molgrid.GridMaker()
gmaker.forward(batch, input_tensor, random_translation=0.0, random_rotation=False)
\end{minted}
\caption{\label{fig:gridmaker} An illustration of \texttt{molgrid.GridMaker} usage, generating a 4-dimensional grid from a batch of molecules, with data layout $NxCxLxWxH$.}
\end{figure}

A \texttt{GridMaker} is used to generate a voxel grid from an \texttt{Example}, an \texttt{ExampleVec}, a \texttt{CoordinateSet}, or paired \texttt{Grid}s of coordinates and types. \texttt{GridMaker} can operate directly on a user-provided Torch tensor or \texttt{Grid}, or it can return into a new NumPy array via \texttt{GridMaker.make\_ndarray} or Torch tensor via \texttt{GridMaker.make\_tensor}. \texttt{GridMaker} features GPU-optimized gridding that will be used if a compatible device is available. \texttt{GridMaker} options pertaining to the properties of the resulting grid are specified when the \texttt{GridMaker} is constructed, while the examples from which a grid will be generated and their instantiation properties (including any transformations) are specified by a particular invocation of \texttt{GridMaker.forward}. Specifically, Listing~\ref{lst:gm_constructor} shows the possible constructor arguments, including the grid resolution; dimension along each side of the cube; whether to constrain atom density values to be a binary indicator of overlapping an atom, rather than the default of a Gaussian to a multiple of the atomic radius (call this $grm$) and then decaying to 0 quadratically at $\frac{1+2grm^{2}}{2grm}$; whether to index the atomic radius array by type id (for vector types); a real-valued pre-multiplier on atomic radii, which can be used to change the size of atoms; and, if using real-valued atomic densities (rather than the alternative binary densities), the multiple of the atomic radius to which the Gaussian component of the density extends. Figure~\ref{fig:gridmaker} shows an example of basic \texttt{GridMaker} usage, default-constructing a GridMaker and using it to populate a grid with densities for a batch of molecules. If desired, the values of random\_translation and random\_rotation can be set in the call to \texttt{gmaker.forward}, thereby applying random data augmentation to each example in the batch.  If it is desirable to retain the applied transformation, then transformations can be created explicitly as shown in Figure~\ref{fig:transform}. The \texttt{GridMaker} class also defines a \texttt{backward} function that computes atomic gradients, which can be used for tasks ranging from visualizing what a network has learned to using a trained network to optimize the coordinates and types of input molecules.   \begin{listing}[]

\begin{minted}[
frame=single,
framesep=2mm,
baselinestretch=1.2,
fontsize=\footnotesize]{python}
gmaker = molgrid.GridMaker(resolution=0.5, dimension=23.5, binary=False, 
radius_type_indexed=False, radius_scale=1.0, gaussian_radius_multiple=1.0)
\end{minted}
\caption{Available arguments to the \texttt{GridMaker} constructor, along with their default values.}
\label{lst:gm_constructor}
\end{listing}

\subsection{Transformations}

\begin{figure}[tbp]
\centering
\includegraphics[width=\linewidth]{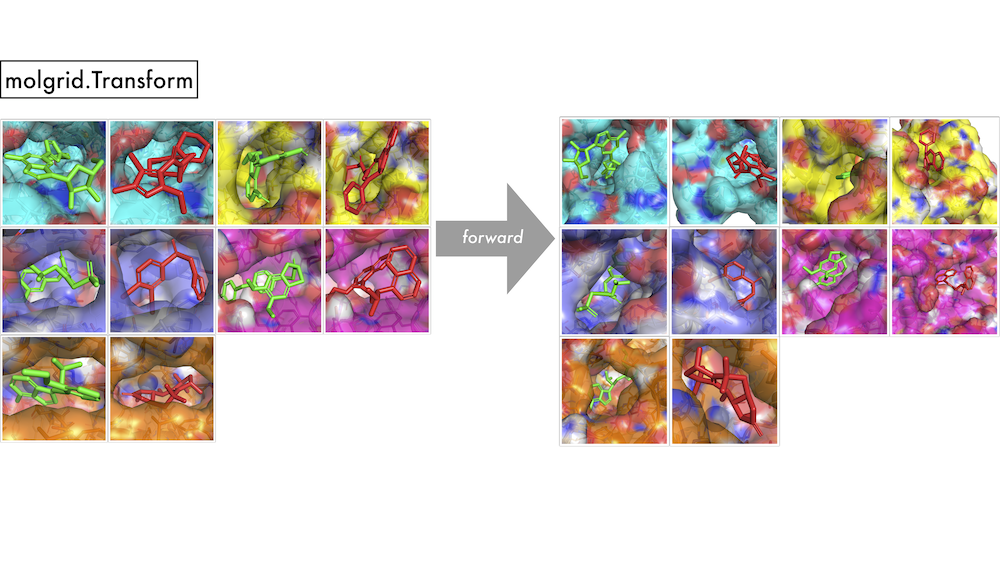}
\begin{minted}[
frame=lines,
framesep=2mm,
baselinestretch=1.2,
fontsize=\footnotesize, linenos]{python}
for data in batch:
    t = molgrid.Transform(center=(0,0,0), random_translate=2.0, random_rotation=True)
    t.forward(data, transformed_data, dotranslate=True)
    # do something with transformed_data
\end{minted}
\caption{\label{fig:transform} An illustration of \texttt{molgrid.Transform} usage, applying a distinct random rotation and translation to each of ten input examples. These transformations can also be applied separately to individual coordinate sets. Transformations to grids being generated via a \texttt{molgrid.GridMaker} can be generated automatically by specifying \texttt{random\_rotation=True} or \texttt{random\_translation=True} when calling \texttt{Gridmaker.Forward}.}
\end{figure}

Data augmentation in the form of random rotations and translations of input examples can be performed by passing the desired options to \texttt{GridMaker.forward} as described in the previous section. Specific translations and rotations can also be applied to arbitrary \texttt{Grid}s,  \texttt{CoordinateSet}s, or \texttt{Example}s by using the \texttt{Transform} class directly. \texttt{Transform}s can store specific rotations, described by a \texttt{libmolgrid::Quaternion}; an origin around which to rotate, described by a \texttt{libmolgrid::float3}, which is also interconvertible with a Python tuple; and a specific translation, expressed in terms of Cartesian coordinates and also described by a \texttt{float3}. Various subsets of these values may also be specified, providing for multiple transformation paradigms; usage examples are shown in Listing~\ref{lst:t_constructor}. These prove useful for sophisticated networks such as the spatial transformer. Figure~\ref{fig:transform} shows the behavior of \texttt{Transform.forward}, taking an input \texttt{Example} and returning a transformed version of that \texttt{Example} in \texttt{transformed\_example}. 

\begin{listing}[]
\begin{minted}[
frame=single,
framesep=2mm,
baselinestretch=1.2,
fontsize=\footnotesize]{python}
# Usage 1: specify a center, maximum distance for random translation, 
and whether to randomly rotate
transform1 = molgrid.Transform(center=molgrid.float3(0.0,0.0,0.0), random_translate=0.0, 
random_rotation=False)

qt = molgrid.Quaternion(1.0, 0.0, 0.0, 0.0)
center = molgrid.float3(0.0, 0.0, 0.0)
translate = molgrid.float3(0.0, 0.0, 0.0)
# Usage 2: specify a particular rotation, to be performed around the molecule's center
transform2 = molgrid.Transform(qt)
# Usage 3: specify a particular rotation and the center around which it will be performed
transform3 = molgrid.Transform(qt, center)
# Usage 4: specify a particular rotation and center, along with a specific translation
transform4 = molgrid.Transform(qt, center, translate)
\end{minted}
\caption{Available \texttt{Transform} constructors.}
\label{lst:t_constructor}
\end{listing}

\section{Results}

\begin{figure}[tbp]
\centering
\begin{subfigure}[t]{.3\linewidth}
\includegraphics[width=\linewidth]{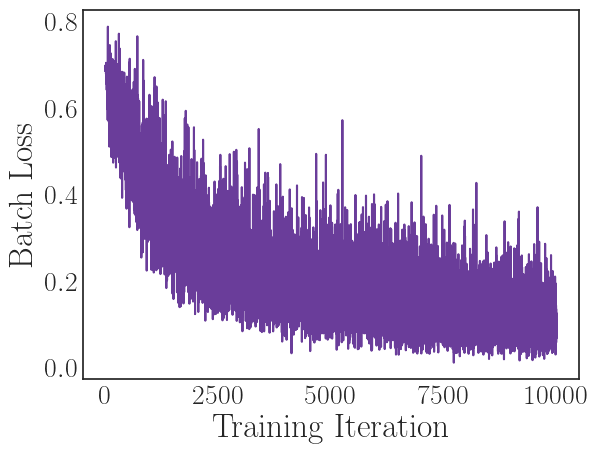}
\caption[]{\label{caffetrain}}
\end{subfigure}%
\hfill
\begin{subfigure}[t]{.3\linewidth}
\includegraphics[width=\linewidth]{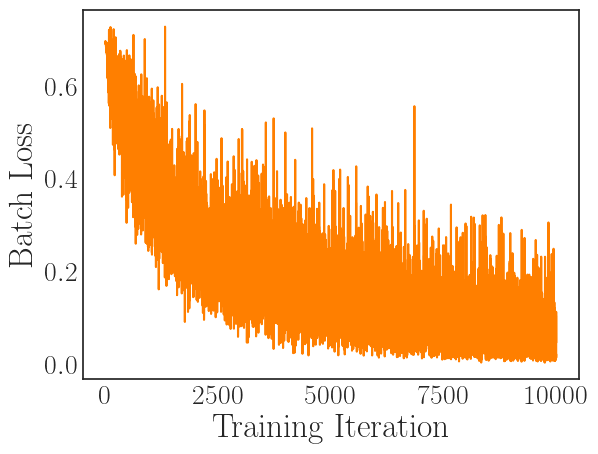}
\caption[]{\label{pytorchtrain}}
\end{subfigure}
\hfill
\begin{subfigure}[t]{.3\linewidth}
\includegraphics[width=\linewidth]{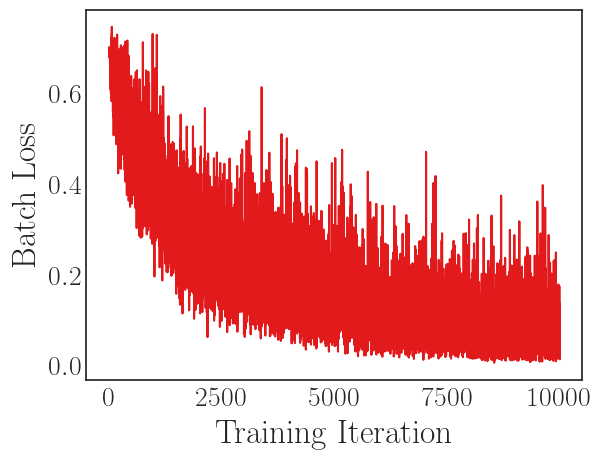}
\caption[]{\label{kerastrain}}
\end{subfigure}\caption{\label{fig:training} Loss per iteration while training a simple model, with input gridding and transformations performed on-the-fly with \texttt{libmolgrid} and neural network implementation performed with \subref{caffetrain} Caffe, \subref{pytorchtrain} PyTorch, and \subref{kerastrain} Keras with a Tensorflow backend.
}
\end{figure}

\begin{figure}[tbp]
\centering
\begin{subfigure}[t]{.3\linewidth}
\includegraphics[width=\linewidth]{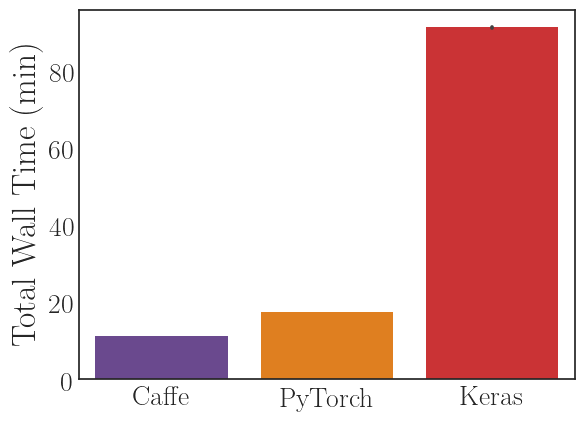}
\caption[]{\label{titanx}}
\end{subfigure}%
\hfill
\begin{subfigure}[t]{.3\linewidth}
\includegraphics[width=\linewidth]{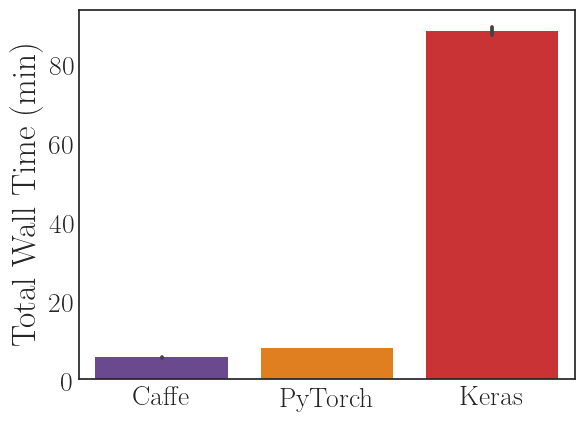}
\caption[]{\label{titanv}}
\end{subfigure}%
\hfill
\begin{subfigure}[t]{.3\linewidth}
\includegraphics[width=\linewidth]{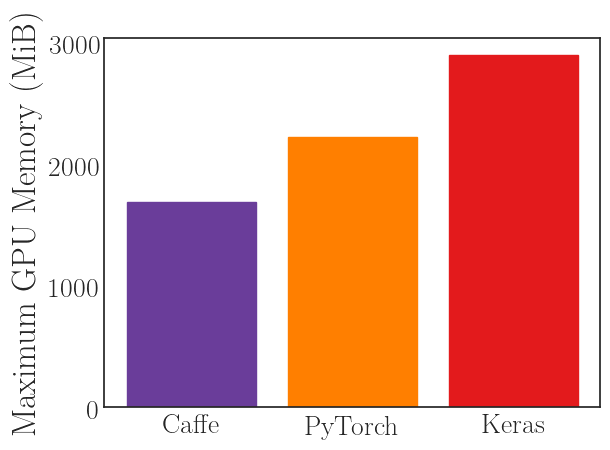}
\caption[]{\label{gpumem}}
\end{subfigure}\caption{\label{fig:genstats} Performance information for using \texttt{libmolgrid} with each major supported neural network library. All error bars are 98\% confidence intervals computed via bootstrap sampling of five independent runs. \subref{titanx} Walltime for training the simple model shown training above using a GTX Titan X. \subref{titanv} Walltime for training the same simple model using a Titan V. \subref{gpumem} Maximum GPU memory utilization while training.
}
\end{figure}

We demonstrate model training with input tensors populated by \texttt{molgrid} and neural networks implemented using Caffe, PyTorch, and Keras with a Tensorflow backend. Figure~\ref{fig:training} shows successful training of a basic feed-forward network on a toy dataset using each of these three deep learning frameworks to perform binary classification of active versus inactive binding modes. Timing was performed using GNU \texttt{time}, while memory utilization was obtained with \texttt{nvidia-smi -q -i 1 -d MEMORY -l 1}. The Caffe data was obtained using \texttt{caffe train} with the model at \url{https://github.com/gnina/models/blob/master/affinity/affinity.model} with the affinity layers removed; the PyTorch data was obtained using \url{https://gnina.github.io/libmolgrid/tutorials/train_basic_CNN_with_PyTorch.html}, run for 10,000 iterations; and the Keras data was obtained using \url{https://gnina.github.io/libmolgrid/tutorials/train_basic_CNN_with_Tensorflow.html}, run for 10,000 iterations. The metadata file for training is at \url{https://github.com/gnina/libmolgrid/blob/master/test/data/small.types}, using structures found at \url{https://github.com/gnina/libmolgrid/tree/master/test/data/structs}. \texttt{molgrid} is fully functional with any of these popular libraries. Its overall speed and memory footprint varies significantly with the user's chosen library, however. As shown in Figure~\ref{fig:genstats}\subref{titanx} and Figure~\ref{fig:genstats}\subref{titanv}, the performance when using a GPU for gridding and neural network training is much faster when using Caffe and PyTorch than it is when using Tensorflow via Keras, with modest improvements in performance for Caffe and PyTorch when using the newer Titan V GPU rather than the older GTX Titan X. This is due to \texttt{libmolgrid}'s ability to directly access underlying data buffers when interoperating with Caffe and PyTorch, thus avoiding unnecessary data migration between the CPU and GPU; this is not currently possible with Tensorflow, and so passes through the network involve grids being generated on the GPU by \texttt{molgrid}, copied into a NumPy array on the CPU, and then copied back onto the GPU by Tensorflow when training begins. This results in a significant performance penalty, with memory transfers fundamentally limiting performance; future versions of \texttt{molgrid} will seek to mitigate this issue with Tensorflow 2.0. The discrepancy in memory utilization shown in Figure~\ref{fig:genstats}\subref{gpumem} is somewhat less dramatic, but similarly, memory utilization when doing neural network training with Tensorflow is less efficient than using the other two libraries. 

\begin{figure}[tbp]
\centering
\begin{subfigure}[t]{.48\linewidth}
\includegraphics[width=\linewidth]{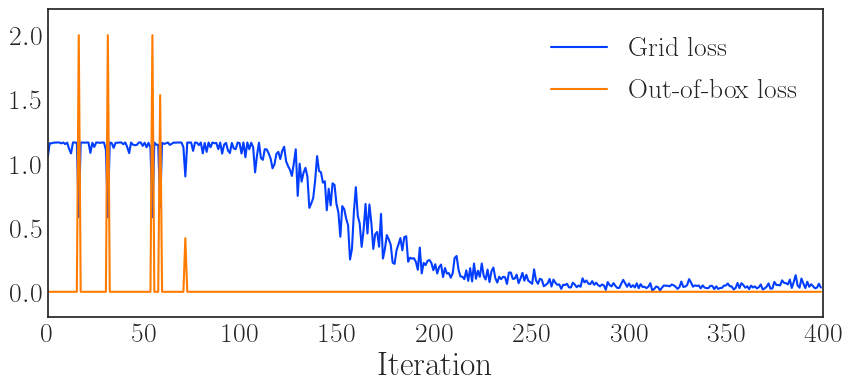}
\caption[]{\label{reduction_training}}
\end{subfigure}%
\hfill
\begin{subfigure}[t]{.48\linewidth}
\includegraphics[width=\linewidth]{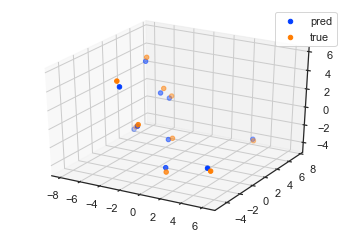}
\caption[]{\label{reduction_result}}
\end{subfigure}\caption{\label{fig:cartred} Cartesian reduction example.
  \subref{reduction_training} Loss per iteration for both the grid loss and
  out-of-box loss for training with naively initialized coordinates, showing \texttt{libmolgrid}'s utility for converting between
  voxelized grids and Cartesian coordinates. \subref{reduction_result} Sampled
  coordinate predictions compared with the true coordinates, demonstrating
  a root mean squared accuracy of 0.09{\AA}.
}
\end{figure}

As an example of a more specialized task that can be performed with libmolgrid, we demonstrate training a CNN to convert voxelized atomic densities to Cartesian coordinates. Each training example consists of a single atom, provided to the network as a voxelized grid for which the network will output Cartesian coordinates. The loss function is a simple mean squared error grid loss for coordinates that fall within the grid, and a hingelike loss for coordinates outside.
As shown in Figure~\ref{fig:cartred}\subref{reduction_training}, the model initially has difficulty learning because the atomic gradients only receive information from the parts of the grid that overlap an atom, but eventually converges to an accuracy significantly better than the grid resolution of 0.5{\AA}.  Example predictions are shown in Figure~\ref{fig:cartred}\subref{reduction_result}. This task could be applicable to a generative modeling workflow, and also demonstrates \texttt{libmolgrid}'s versatility as a molecular modeling tool. 

\section{Conclusion}

Machine learning is a major research area within computational chemistry and drug discovery, and grid-based representation methods have been applied to many fundamental problems with great success. No standard library exists for automatically generating voxel grids or tensor representations more generally from molecular data, or for performing the basic tasks such as data augmentation that typically must be done to achieve high predictive capability on chemical datasets using these methods. This means that researchers hoping to pursue methodological advances using grid-based methods must reproduce the work of other groups and waste time with redundant programming. \texttt{libmolgrid} attempts to reduce the amount of irrelevant work researchers must do when pursuing advances in grid-based machine learning for molecular data, by providing an efficient, concise, and natural C++/CUDA and Python API for data resampling, grid generation, and data augmentation. It also supports spatial and temporal recurrences over input, allowing for size extensiveness even while using cubic grids (by performing a subgrid decomposition), and processing of simulation data such as molecular dynamics trajectories while preserving temporal ordering of frames, if desired. With adoption, it will also help standardize performance, enhance reproducibility, and facilitate experimentation among computational chemists interested in machine learning methods. \texttt{libmolgrid} support for Caffe and PyTorch is complete, while we plan to enhance Tensorflow support by taking advantage of the Tensorflow 2.0 programming model and avoiding the unnecessary data transfers that currently limit combined \texttt{libmolgrid}-Tensorflow performance. Other future enhancements will include the ability to generate other types of grids, for example spherical ones. Documentation, tutorials, and installation instructions are available at \url{http://gnina.github.io/libmolgrid}, while the source code along with active support can be found at \url{https://github.com/gnina/libmolgrid}. 

\bibliography{main}

\end{document}